\pdfoutput=1

\documentclass[11pt]{article}

\usepackage[]{acl}

\usepackage{times}
\usepackage{latexsym}

\usepackage[T1]{fontenc}

\usepackage[utf8]{inputenc}

\usepackage{microtype}
\usepackage{xcolor}
\usepackage{color}
\usepackage{amsmath}
\usepackage{bbm}
\usepackage{algorithm}
\usepackage{algorithmic}
\usepackage{amsmath}
\usepackage{bbm}
\usepackage{adjustbox}
\usepackage{bm}
\usepackage{hyperref}
\usepackage{amsmath}
\usepackage{multirow}
\usepackage{diagbox}
\usepackage{booktabs,siunitx}
\usepackage{amssymb}

\title{Exploiting Unlabeled Data for Target-Oriented Opinion Words Extraction}

\author{
	Yidong Wang\textsuperscript{\normalfont 1 $^*$} \quad 
	Hao Wu\textsuperscript{\normalfont 1 \thanks{\enspace\ Equal Contribution.}} \quad 
	Ao Liu\textsuperscript{\normalfont 1} \quad 
	Wenxin Hou\textsuperscript{\normalfont 2} \\
	\textbf{Zhen Wu}\textsuperscript{\normalfont 3 \thanks{$^\dagger$ Corresponding author.}} \quad 
	\textbf{Jindong Wang}\textsuperscript{\normalfont 4} \quad 
	\textbf{Takahiro Shinozaki}\textsuperscript{\normalfont 1} \quad 
	\textbf{Manabu Okumura}\textsuperscript{\normalfont 1}
	\quad 
	\textbf{Yue Zhang}\textsuperscript{\normalfont 5}
	\\ 
 	\textsuperscript{1}Tokyo Institute of Technology \quad
 	\textsuperscript{2}Microsoft STCA \\
 	\textsuperscript{3}Nanjing University \quad 
 	\textsuperscript{4}Microsoft Research Asia \quad 
 	\textsuperscript{5}Westlake University
 	\\ \texttt{\{yidongwang37, wu.364371691\}@gmail.com}, \texttt{wuz@nju.edu.cn}
}	

\begin{document}
\maketitle
\begin{abstract}
    Target-oriented Opinion Words Extraction (TOWE) is a fine-grained sentiment analysis task that aims to extract the corresponding opinion words of a given opinion target from the sentence. Recently, deep learning approaches have made remarkable progress on this task. Nevertheless, the TOWE task still suffers from the scarcity of training data due to the expensive data annotation process. Limited labeled data increase the risk of distribution shift between test data and training data. In this paper, we propose exploiting massive unlabeled data to reduce the risk by increasing the exposure of the model to varying distribution shifts. Specifically, we propose a novel Multi-Grained Consistency Regularization (MGCR) method to make use of unlabeled data and design two filters specifically for TOWE to filter noisy data at different granularity. Extensive experimental results on four TOWE benchmark datasets indicate the superiority of MGCR compared with current state-of-the-art methods. The in-depth analysis also demonstrates the effectiveness of the different-granularity filters. Our codes are available at \url{https://github.com/TOWESSL/TOWESSL}.
\end{abstract}

\section{Introduction}

\begin{figure}[!t]
	\centering
	\includegraphics[width=\linewidth]{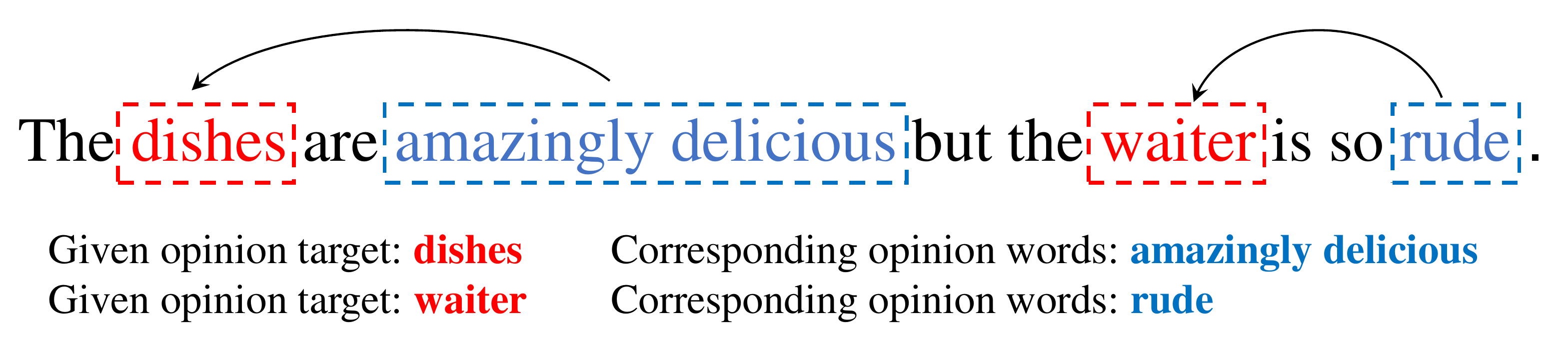}
	\caption{Example of TOWE. Words in red are opinion targets and words in blue are corresponding opinion words. TOWE extracts corresponding opinion words when given opinion targets.  }
	\label{fig:task}
\end{figure}

Target-oriented Opinion Words Extraction (TOWE)~\cite{fan2019towe} is an important subtask of aspect-based sentiment analysis (ABSA)~\cite{semEval14}, which aims to extract the corresponding opinion words for a given opinion target from the sentence. For the TOWE task, opinion targets, also called aspect terms, are the entities or objects in the sentence toward which users show attitudes; opinion words, sometimes known as opinion expressions, are those words explicitly mentioned in the sentence and used to express attitudes or opinions. Figure \ref{fig:task} shows an example of the TOWE task. For the sentence ``\textit{The dishes are amazingly delicious but the waiter is so rude.}'', the terms ``\textit{dishes}'' and ``\textit{waiter}'' are two opinion targets. The goal of TOWE is to extract ``\textit{amazingly delicious}'' as the opinion words for the opinion target ``\textit{dishes}'' and opinion word ``\textit{rude}'' when given the opinion target ``\textit{waiter}''.

While seminal work casts TOWE as a sequence labeling problem, using fully supervised learning methods to identify opinion words and phrases from sentences, recent work shows that external sources of information can be highly useful for improving the performance. In particular, both syntactic knowledge \cite{dai2022reasoning,veyseh2020introducing,jiang2021attention,zhang2021enhancement} and sentiment information \cite{aaai/lotn} have been exploited, with the former helping to identify the correlation between opinion targets and opinion expressions, and the latter helping to identify words and phrases that are correlated with sentiment polarities. Existing work integrates these external features via representation structures \cite{dai2022reasoning,veyseh2020introducing,jiang2021attention} and multi-task learning \cite{aaai/lotn,zhang2021enhancement}.

Intuitively, the goal of opinion words extraction is to obtain structured knowledge from raw data, and therefore ideally the amount of test data, namely the data from which opinion words are mined, should be large. This can inevitably increase the risk of distribution shift between test instances and the training data, even if the test data is from the same domain \cite{ganin2016domain}. Existing work, however, adopt a supervised training setting, with the model being tuned on a set of fixed training data. To address this issue, we consider making use of raw text inputs to increase the exposure of the model to varying distribution shifts. Our main idea is to (1) find raw data in a similar domain as the target-oriented opinion words extraction training data (i.e. gold data), (2) using a model trained on the gold data to assign silver labels, and then (3) design a set of filters to select the most useful set of silver data, so that (4) the selected silver data can be used together with the gold data for training a final model.

As a pre-processing step, we train a opinion target extraction model, which is used to label the set of raw data. Those raw sentences without any opinion target are filtered. The resulting raw data has the same format as the input structure of TOWE data. Then in our implementation, steps (2) to (4) above are done in a joint batch training process. First, we initialize a TOWE model using BERT. Then in each batch, we randomly sample a subset of gold data and a subset of raw data, using the current model to assign pseudo labels to the raw data. According to the current model probabilities and an external sentiment classifier, we filter the silver data by removing low-confidence sentences, as well as masking low-confidence words from the remaining sentences. Finally, the model parameters are updated by using the standard cross-entropy losses separately on both the gold and silver data, so that the next batch of training can start with the new model. The training process continues for a fixed number of iterations, and the model with the highest development scores are selected for testing.

Results on four standard benchmarks show that our method gives significantly improved results when raw data are used. We achieve the best reported results on all the datasets. In addition, the critical ablation demonstrates our sentence-level and word-level noises filtering both bring improvements for the final results. In-depth analysis shows our method significantly reduces the extraction errors on different error types. To our knowledge, we are the first to consider the use of unlabeled data for TOWE, successfully giving state-of-the-art results on benchmarks. Our code and datasets are available at \url{https://github.com/TOWESSL/TOWESSL}.

\begin{figure}[!htbp]
    \centering
    \includegraphics[width=\linewidth]{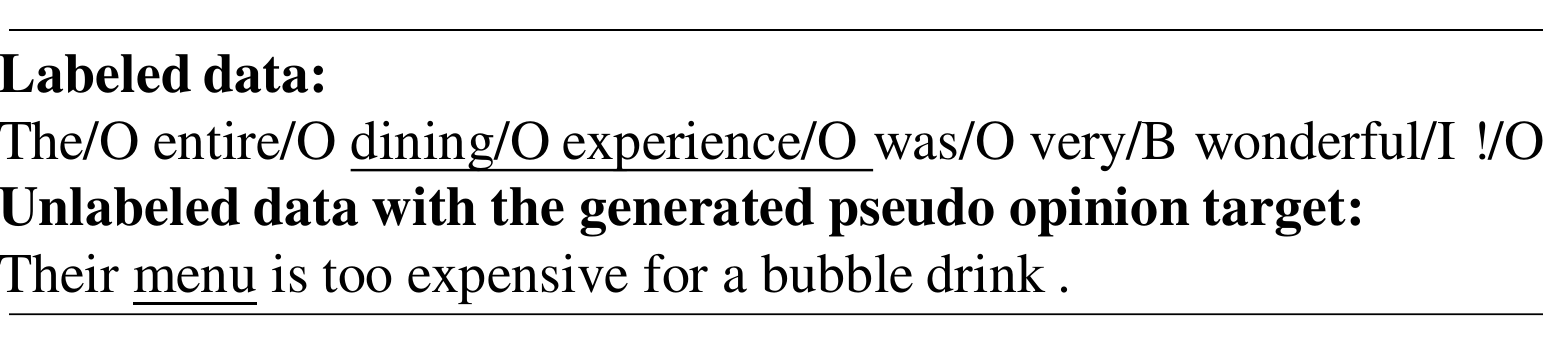}
    \caption{Examples of labeled data and unlabeled data with the pseudo opinion target. Words with underline indicate opinion targets. The span in the labeled data beginning with $B$ and followed by $I$ represent the corresponding opinion words.}
    \label{fig:example1}
\end{figure}

\begin{figure*}[!htbp]
    \centering
    \includegraphics[width=0.8\linewidth]{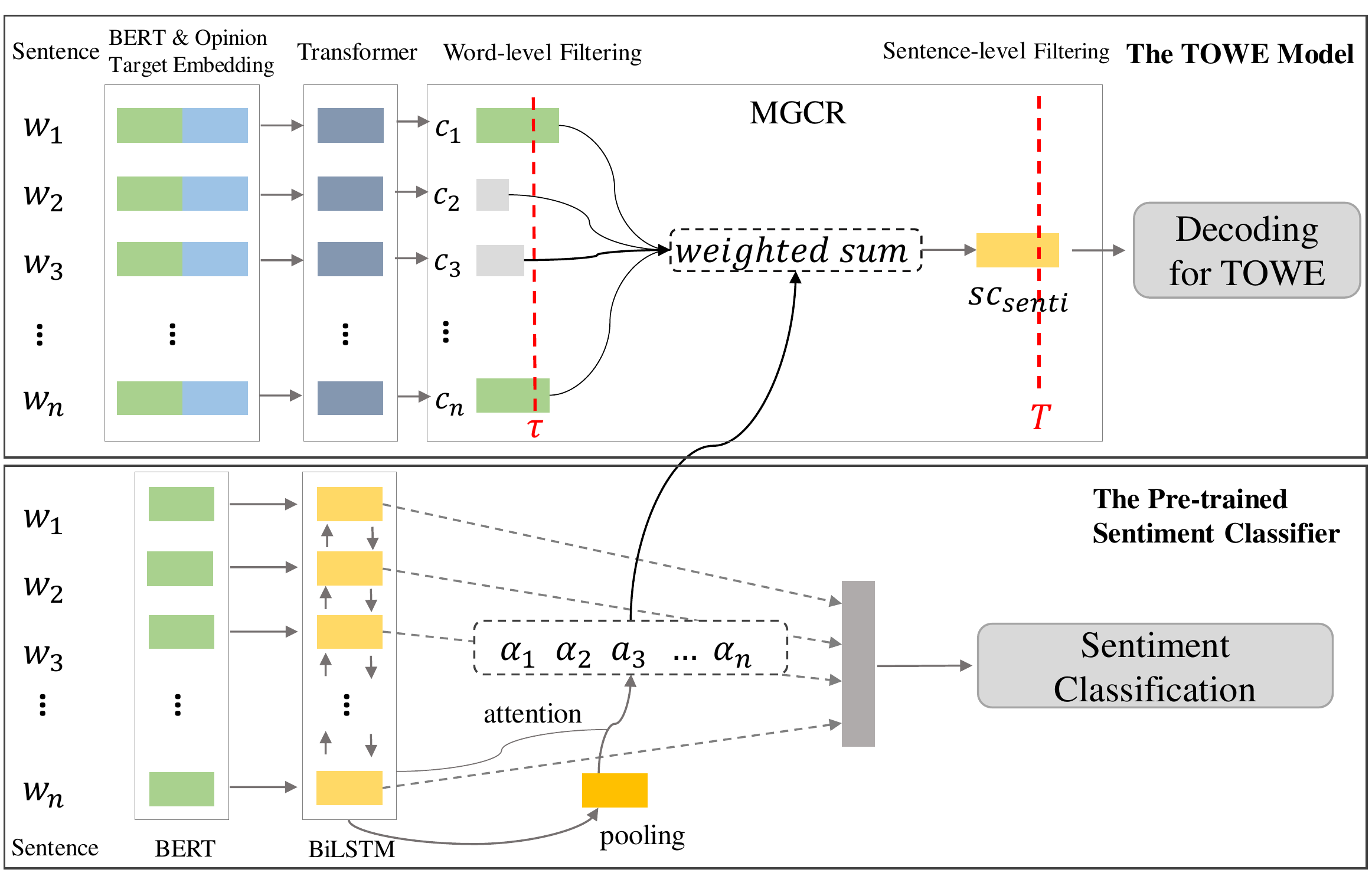}
    \caption{Overview of the architecture of multi-grained consistency regularization. For simplicity, we mark the confidence of $i$-th word as $c_i$. Note that the input sentence of the TOWE model is the same as the input sentence of the pre-trained sentiment classifier.}
    \label{fig:MGCR}
\end{figure*}

\section{Background}
\subsection{Task Formalization}
The TOWE task requires the opinion target as the input and extracts the target-oriented opinion words from the sentence. It can be formulated as a target-oriented sequence labeling task. In this work, we use the notation $s$ to represent a labeled sentence from the TOWE dataset, and use $s^u$ to denote a unlabeled sentence. Formally, given an input sentence $s=\{w_1,w_2,\dots,w_n\}$ consisting of $n$ words and an opinion target $w_t$ in $s$ (here we notate an opinion target as one word for simplicity and $t$ is the position of the opinion target in the sentence), the goal of TOWE is assign a corresponding label $y_i \in\{B,I,O\}$($B$: \textbf{B}eginning, $I$: \textbf{I}nside, $O$: \textbf{O}thers) for each word $w_i$ in $s$. The spans beginning with $B$ and followed by $I$ represent the corresponding opinion words of the opinion target $w_t$.
Figure \ref{fig:example1} shows an sequence labeling example of TOWE.


\subsection{Pseudo Opinion Targets Labeling}
 Raw unlabeled data can not be directly used in the semi-supervised scenario for TOWE, as they lack the necessary annotations of opinion targets. We use the opinion targets in labeled TOWE data as the ground truth to train a BERT-based target extraction model. 

Specifically, given an input sentence $s=\{w_1,w_2,\dots,w_n\}$ from the TOWE dataset, we employ a BERT model~\cite{devlin2019bert} to generate the context representation of the each word $w_i$ as follows:
\begin{equation}
    {\mathbf{h}_1^{pt}, \dots, \mathbf{h}_n^{pt}}=\text{BERT}(w_1, \dots, w_n)
\end{equation}

Then the context representation $\mathbf{h}_i^{pt}$ of the word $w_i$ is fed to a linear layer and a softmax layer to predict the corresponding label. Similar to TOWE, the BIO scheme is used. We train the opinion target extraction model by minimizing the cross-entropy loss between the predicted BIO label distribution and the ground truth. After training, we can employ the opinion target extraction model to obtain the pseudo opinion targets of the unlabeled sentence $s^u$, which makes semi-supervised learning feasible for the TOWE task.

 
\subsection{Consistency Regularization}
Consistency regularization is a semi-supervised learning method that shows effectiveness in a variety of fields~\cite{sohn2020fixmatch,chen2020mixtext,xie2020unsupervised,zhang2021flexmatch}. It improves the generalization performance of a TOWE model by generating a perturbed version $\omega(s^u)$ of the original unlabeled sentence $s^u$ and forcing the predicted category of each word to be the same, where $\omega$ is the perturbing function~\cite{lee2013pseudo}.

To perform semi-supervised learning for TOWE, we feed the unlabeled sentence $s^u$ and the position $t$ of the pseudo opinion target $w^u_t$ to the TOWE model (see section~\ref{towe_model}), and obtain the predicted probability $p_i(y|\theta;s^u, t)$ of the word $w_i^u$, where $\theta$ represents the parameters of the TOWE model. A vanilla consistency loss for consistency regularization of TOWE is computed as:
\begin{equation}
\label{eq-vanilla_con}
    \begin{split}
        \frac{1}{n}\sum_{i=1}^{n} \mathcal{H}( \hat{p}_i(y|\theta;s^u, t), p_i(y|\theta;\omega(s^u), t) ),
    \end{split}
\end{equation}
where $\hat{p}_i(y|\theta;s^u, t)=\arg\max p_i(y|\theta;s^u, t)$ and $\hat{p}_i(y|\theta;s^u, t)$ denotes the predicted label of the $w_i^u$, the $\mathcal{H}(\cdot, \cdot)$ refers to the cross-entropy loss.
In this work, we use Random Mask and Random Synonym Replacement by using WordNet~\cite{miller1995wordnet} as the perturbing function $\omega$.

\section{Method}

Figure \ref{fig:MGCR} shows the the framework of our \textit{Multi-Grained Consistency Regularization (MGCR)} method. The TOWE model is a BERT-based neural sequence labeling network. As mentioned in the introduction, we initialize a BERT TOWE model (Section~\ref{towe_model}), and then iteratively conduct batch training. In each batch, we impose a standard cross-entropy loss on a set of sampled gold data (Eq~\ref{eq-sup}), and a regularization loss on a set of randomly sampled raw data with targets (Eq~\ref{eq-final_con}). The latter is defined by using the current model to assign TOWE labels, and then adding a sentence-level (Section~\ref{eq-sent_conf}) and a word-level (Section~\ref{eq-word_conf}) filter. Besides, as mentioned in the introduction, MGCR exploits latent opinion words from a pre-trained review sentiment classification~\cite{aaai/lotn} to filter noises more accurately.

\subsection{TOWE Model}
\label{towe_model}
For the sentence $s=\{w_1,w_2,\dots,w_n\}$, the TOWE model first employs a pre-trained BERT to generate the context representations $\{\mathbf{h}_1,\dots,\mathbf{h}_n\}$ of $s$. To incorporate the opinion target information into the sentence representations, we use the number $1$ to represent the position of opinion target $w_t$ in $s$ and $0$ to denote the positions of other words, and then map them to the position embedding sequence $\{\mathbf{e}_1,\dots,\mathbf{e}_n\}$. We integrate opinion target information by concatenating, i.e., $\mathbf{\widetilde{h}}_i = [\mathbf{h}_i; \mathbf{e}_i]$.

To better encode opinion target information, we additionally employ a multi-layer Transformer architecture~\cite{vaswani2017attention} to generate the target-specific context representations as follows:
\begin{equation}
    \mathbf{r}_1, \dots, \mathbf{r}_n=\text{Transformer}(\mathbf{\widetilde{h}}_1, \dots, \mathbf{\widetilde{h}}_n).
\end{equation}

Finally, we use the representation $\mathbf{r}_i$ to predict the opinion word probability $p_i(y|\theta;s, t)$ of the word $w_i$ when given the opinion target $w_t$:
\begin{equation}
    p_i(y|\theta;s, t)=\text{softmax}(\mathbf{W}_r\mathbf{r}_i+\mathbf{b}_r),
\end{equation}
where $\mathbf{W}_r$ and $\mathbf{b}_r$ are learnable weight and bias.
The loss of supervised learning on the TOWE dataset is defined as:
\begin{equation}
    \label{eq-sup}
    \mathcal{L}_{s} = \frac{1}{n}\sum_{i=1}^{n} \mathcal{H}(y_i,p_i(y|\theta;s, t)).
\end{equation}

\subsection{Multi-Grained Consistency Regularization}

\subsubsection{Sentence-level Filtering}
Given an unlabeled input sentence ${s^u}=\{{w}^u_1,{w}^u_2,\dots,{w}^u_n \}$ with $w^u_t$ as the pseudo opinion target position, the sentence-level confidence $sc_{avg}$ is defined as follows: 
\begin{equation}
\label{eq-van_senti_sen_conf}
    \begin{split}
         & sc_{avg} =  \frac{1}{n}\sum_{i=1}^n  \max (p_i (y|\theta; s^u,t)), \\
    \end{split}
\end{equation}
where $\max (p_i (y|\theta; s^u,t))$ (i.e., the maximum of the probabilities of $B$, $I$, and $O$) is the confidence of the $i$-th word.
Sentences with confidences below the given threshold $T$ are masked during training. 
In Eq.~\eqref{eq-van_senti_sen_conf}, different words in the sentence are treated equally, which ignores the importance of the opinion words. 
We highlight the confidences of opinion words identified by their larger sentiment scores.
Specifically, we obtain the sentiment-attention scores from a pre-trained attention-based sentiment classifier parameterized by $\theta_{senti}$ from \cite{aaai/lotn}\footnote{The sentiment classifier we use is the same as \cite{aaai/lotn}, except that we use BERT as the word representation model while they use GloVe vectors~\cite{pennington2014glove}.}. After acquiring the representations $\{{z}_1,{z}_2, \dots, {z}_n \}$ of the unlabeled sentence, the attention score $\alpha_i$ of ${z}_i$ is calculated as follows:
\begin{equation}
\label{eq-senti_attn}
    \begin{split}
         & {z}_{avg} = \frac{1}{n} \sum_{i=1}^n {z}_i, \\
         & f({z}_i,{z}_{avg})= {z}_i \cdot {\mathbf{W}} \cdot {z}_{avg} + {\mathbf{b}},\\
         & \alpha_i = \frac{e^{f({z}_i,{z}_{avg})}}{\sum_{j=1}^n e^{f({z}_j,{z}_{avg})}},
    \end{split}
\end{equation}
where $\mathbf{W}$ and $\mathbf{b}$ are learnable weight and bias. We then compute the sentiment-aware confidence based on the obtained attention scores as:
\begin{equation}
    \label{eq:sc_senti}
    sc_{senti} =  \sum_{i=1}^n \alpha_i \cdot \max (p_i (y|\theta; s^u,t)).
\end{equation}

Similar to Eq.~\eqref{eq-van_senti_sen_conf}, any sentence whose sentiment-aware confidence is lower than the threshold $T$ will be masked in consistency regularization. The final filtering mechanism of noisy sentences can be expressed as:
\begin{equation}
    \label{eq-sent_conf}
    \begin{split}
        \mathbbm{1}(sc_{senti}>T),
    \end{split}
\end{equation}
where $\mathbbm{1}(\cdot > T)$ is the indicator function for confidence thresholding with $T$ being the threshold.

\subsection{Word-level Filtering}
To further reduce the noise in the unlabeled data,
we filter the noisy words with a more fine-grained confidence thresholding mechanism. The filtering of noisy words can be expressed as
\begin{equation}
    \label{eq-word_conf}
    \begin{split}
        \mathbbm{1}(\max(p_i(y|\theta;s^u, t))>\tau).
    \end{split}
\end{equation}

Any word with a confidence lower than $\tau$ is masked. Note that the word-level threshold $\tau$ can be different from the sentence-level threshold $T$.

\subsection{Training Objective} For labeled sentences, the supervised loss is the same as Eq.~\eqref{eq-sup}. For unlabeled sentences, the consistency loss is:
\begin{equation}
\label{eq-final_con}
    \begin{split}
        \mathcal{L}_c=&\mathbbm{1}(sc_{senti}>T)
        \\&\cdot\{\frac{1}{n}\sum_{i=1}^{n} \mathbbm{1}(\max(p_i(y|\theta;s^u, t))>\tau)
        \\&\cdot \mathcal{H}( \hat{p}_i(y|\theta;s^u, t), p_i(y|\theta;\omega(s^u), t) )\}.
    \end{split}
\end{equation}

The final training objective is given by:
\begin{equation}
    \mathcal{L} = \mathcal{L}_{s} + \mathcal{L}_{c}.
\end{equation}

\section{Experiments}

\begin{table}[!t]
	\small
	\centering

	 \scalebox{1.0}{
	\begin{tabular}{cc|cc}
		\midrule
		\multicolumn{2}{c|}{Datasets} & {\#sentences} & {\#opinion targets}\\
		\midrule
		\multirow{2}*{14res}  & Train & 1,627 & 2,643 \\
		& Test & 500 & 864\\
		\midrule
		\multirow{2}*{15res}  & Train & 754 & 1,076\\
		& Test & 325 & 436 \\
		\midrule
		\multirow{2}*{16res}  & Train & 1,079 & 1,512\\
		& Test & 329 & 457 \\
		\midrule
		\multirow{2}*{14lap}  & Train & 1,158 & 1,634\\
		& Test & 343 & 482 \\
		\midrule		
		Yelp & Unlabeled & 100,000 & -\\
		\midrule		
		Amazon & Unlabeled & 100,000 & -\\
		\midrule
	\end{tabular}
	}
	\caption{Statistics of TOWE datasets and unlabeled datasets. For TOWE datasets, sentence may contain multiple opinion targets. For unlabeled datasets, we randomly sampled data from Yelp for 14res, 15res, 16res datasets and Amazon for 14lap dataset. The unlabeled data is available at \url{https://github.com/TOWESSL/TOWESSL}. }
		\label{tab:dataset}
	
\end{table}

\subsection{Datasets and Metrics}
Following previous studies~\cite{fan2019towe, aaai/lotn, veyseh2020introducing,mao2021joint,jiang2021attention,feng2021target,zhang2021enhancement}, we conduct evaluations on four benchmark datasets for TOWE. The suffixes `res' and `lap' refer to restaurant reviews and laptop reviews, respectively. For the 14res, 15res, and 16res datasets, we use the unlabeled sentences from Yelp\footnote{\url{https://www.yelp.com}}. For the 14lap dataset, we use the unlabeled sentences from Amazon Electronics\footnote{\url{https://www.amazon.com}}~\cite{ni2019justifying}. The statistics of these datasets are listed in Table \ref{tab:dataset}.

We use the evaluation metrics of precision, recall, and F1 score to measure the performance of different methods following previous studies~\cite{fan2019towe, aaai/lotn,jiang2021attention}. An extraction is considered correct only if opinion words from the beginning to the end are all correctly predicted.

\subsection{Experimental Settings}
For our MGCR method, we set the hidden size of both the BERT and the Transformer to 512. The mini-batch sizes of labeled and unlabeled data are set to 16 and 96, respectively. All parameters are optimized using the AdamW optimizer~\cite{loshchilov2018fixing} with an initial learning rate 2e-5 for BERT and 2e-4 for others. We randomly split 20\% of the training set as the validation set and used early stopping. We search different combinations of sentence-level and word-level confidence thresholds for each dataset and use the ones with best validation performances. The experimental results of different threshold combinations are provided in Table~\ref{tbl:thr-14res}, Table~\ref{tbl:thr-15res}, Table~\ref{tbl:thr-16res} and Table~\ref{tbl:thr-14lap}. The training details of the pseudo opinion targets generator and the sentiment classifier are provided in Table~\ref{tbl:para}, respectively. We pre-train the sentiment classifier on the same data as the unlabeled data used for MGCR.

\begin{table}[t!]
\resizebox{\linewidth}{!}{
\begin{tabular}{l|c|c}
\midrule
Hyperparameter & TOWE model & Sentiment Classifier \\ \midrule
Batch size &16(96) & 128 \\
Epochs &50 & - \\
Steps &- & 3000 \\
Learning rate (BERT) &2e-5 & 1e-5 \\
Learning rate (Others) &2e-4 & 1e-4 \\
Hidden dimension &512&  512 \\
Optimizer &AdamW & AdamW \\ \midrule
\end{tabular}
}
\caption{Experimental setting of the training of the TOWE model and the sentiment classifier. For the TOWE model, batch size for labeled data is 16 and 96 for unlabeled data.}
\label{tbl:para}
\centering
\end{table}

\subsection{Baselines}
We compare our MGCR method with the following methods.

\begin{table*}[!htbp]
	\resizebox{\textwidth}{!}{
		\begin{tabular}{l|c c c|c c c|c c c|c c c}
			\midrule
			\multirow{2}{*}{Methods} & \multicolumn{3}{c|}{14res} & \multicolumn{3}{c|}{15res} & \multicolumn{3}{c|}{16res} & \multicolumn{3}{c}{14lap}\\ 
			 
			\cline{2-13}
			&P& R& F1 &P& R& F1&P& R& F1&P& R& F1 \\
			\midrule
			
			Distance-rule~\cite{fan2019towe} & 58.39&43.59&49.92 & 54.12&39.96&45.97 & 61.90&44.57&51.83 & 50.13&33.86&40.42\\ 
			
			Dependency-rule~\cite{fan2019towe} & 64.57&52.72&58.04 &  65.49&48.88&55.98 & 76.03&56.19&64.62& 45.09&31.57&37.14\\
			
			
			TC-BiLSTM~\cite{fan2019towe}& 67.65&67.67&67.61 & 66.06&60.16&62.94 & 73.46&72.88&73.10  & 62.45&60.14&61.21\\ 
			
			IOG~\cite{fan2019towe}& 82.85 & 77.38 & 80.02 & 73.24 & 69.63 & 71.35 & 76.06 & 70.71 & 73.25 & 85.25 & 78.51 & 81.69\\ 
			
			LOTN~\cite{aaai/lotn} & 84.00 & 80.52& 82.21 &  76.61 &  70.29 &  73.29 &  86.57 &  80.89 &  83.62 &  77.08 & 67.62&  72.02\\
			
  			
  			ONG~\cite{veyseh2020introducing} & 83.23&81.46&82.33&76.63&81.14&78.81&87.72&84.38&86.01&73.87&77.78&75.77\\
  			Dual-MRC~\cite{mao2021joint} & \textbf{89.79}&78.43&83.73&77.19&71.98&74.50&86.07&80.77&83.33&78.21&\textbf{81.66}&79.90\\
  		    PER~\cite{dai2022reasoning} & 86.43 &80.39 &83.30 &81.50&75.05&78.14&90.00&84.00&86.90 &80.68&70.72&75.38 \\
  			ARGCN~\cite{jiang2021attention} & 87.32&83.59&85.42&78.81&77.69&78.24&88.49&84.95&86.69&75.83&76.90&76.36\\
  			TSMSA~\cite{feng2021target} & -&-&86.37&-&-&81.64&-&-&89.20&-&-&\underline{82.18}\\
  			MRC-MVT~\cite{zhang2021enhancement} & 86.31&\textbf{89.42}&87.83&82.04&\underline{81.54}&81.79&\underline{90.60}&88.19&89.38&79.59&81.12&80.84\\
  			\midrule
  			MGCR (ours) & \underline{88.65}&\underline{89.36}&$\textbf{89.01}^{\dagger}$ & \underline{84.29}&\underline{83.37}&$\textbf{83.80}^{\dagger} $ &\textbf{91.31}&\textbf{91.74}& $\textbf{91.51}^{\dagger} $ &\textbf{83.76}&\underline{81.25}& $\textbf{82.47}^{\dagger}$\\
  			
  			\midrule
	 \end{tabular} }
	 	\caption{Main results (\%) including recall, precision and F1-score. The best results are in bold and second-best results are underlined. Results of all comparison methods were copied from the original papers. The marker${\  }^{\dagger}$ represents that MGCR outperforms other methods significantly ($p < 0.01$) .} \label{tab-mainresults}
	\centering
\end{table*}

\begin{table*}[!htbp]
	\resizebox{\textwidth}{!}{
		\begin{tabular}{l|c c c|c c c|c c c|c c c}
			\midrule
			\multirow{2}{*}{Methods} & \multicolumn{3}{c|}{14res} & \multicolumn{3}{c|}{15res} & \multicolumn{3}{c|}{16res} & \multicolumn{3}{c}{14lap} \\ 
			\cline{2-13}
			&P& R& F1 &P& R& F1&P& R& F1&P& R& F1 \\
			\midrule
			MGCR & 88.65 & \textbf{89.36} & \textbf{89.01} & \textbf{84.29} & 83.37 & \textbf{83.80} & \textbf{91.31} &\textbf{91.74} & \textbf{91.51} & 83.76 & \textbf{81.25} & \textbf{82.47} \\
			
			\textbf{w/o} Pre-trained Sentiment Classifier & 87.69 & 89.03 & 88.35 & 82.79 & 82.89 & 82.77 & 90.67 & 90.60 & 90.63 & \textbf{84.18} & 79.19 & 81.05 \\
			
			\textbf{w/o} Filtering Noisy Unlabeled Sentences & \textbf{88.84} & 88.00 & 88.41 & 80.13 & \textbf{85.39} & 82.62 & 89.59 & 91.68 & 90.62 & 82.84 & 78.83 & 80.77 \\

			\textbf{w/o} Filtering Noisy Unlabeled Words & 87.29 & 88.12 & 87.70 & 80.10 & 85.33 & 82.66 & 91.02 & 91.30 & 91.16 & 81.99 & 80.19 & 81.07 \\
			
			\textbf{w/o} Consistency Regularization (Labeled Data Only) & 87.34 & 87.05 & 87.19 & 82.42 & 81.81 & 82.11 & 87.19 & 88.38 & 87.76 & 81.70 & 77.89 & 79.70 \\
		
  			\midrule
	 \end{tabular}}
	 \caption{Ablation study results (\%) when removing different components from MGCR method.} 	\label{tbl:abl}
	\centering
\end{table*}

\textbf{Distance-rule}~\cite{fan2019towe} uses  POS tags and regards the nearest adjective to the opinion target as the corresponding opinion word.

\textbf{Dependency-rule}~\cite{fan2019towe} builds templates from the training set with the POS tags of opinion targets and opinion words and the dependency path between them, then uses the hign-frequency dependency templates for target-oriented opinion words extraction on the testing set.

\textbf{TC-BiLSTM}~\cite{fan2019towe} follows the design of the work for target-oriented sentiment classification~\cite{tang2016effective} and concatenate an opinion target embedding for each word position to perform sequence labeling.

\textbf{IOG}~\cite{fan2019towe} employs six different positional and directional LSTMs to encode sentence and then extract the opinion words of the target.

\textbf{LOTN}~\cite{aaai/lotn} transfers the latent opinion knowledge from sentiment classification task into the TOWE task via an auxiliary learning task.

\textbf{ONG}~\cite{veyseh2020introducing} leverages syntax-based opinion possibility scores and the syntactic connections between the words for TOWE.

\textbf{Dual-MRC}~\cite{mao2021joint} used BERT as the encoder and  transforms the TOWE task into a question answering (QA) problem to solve.

\textbf{PER}~\cite{dai2022reasoning} proposes a padding-enhanced reinforcement learning model on both sequential structure and syntactic structure to extract opinion words for opinion targets.

\textbf{ARGCN}~\cite{jiang2021attention} proposes a directed syntactic dependency graph and a attention-based relational graph convolutional neural network to exploit syntactic information for TOWE.

\textbf{TSMSA}~\cite{feng2021target} design a target-specified sequence labeling with multi-head self-attention based on transformer architecture.

\textbf{MRC-MVT}~\cite{zhang2021enhancement} leverages a machine reading comprehension model trained with a multiview paradigm to extract target-oriented opinion words.


\subsection{Main Results and Discussion}
Table \ref{tab-mainresults} shows main results of different methods on four benchmarks. These results demonstrate that MGCR achieves the best F1-score on all datasets. Concretely, MGCR has the following advantages:

\begin{itemize}
\item MGCR significantly outperforms other methods. In these methods, rule-based methods (Distance-rule and Dependency-rule) achieve the worst performance, since they lack robustness and only cover a small number of cases. By contrast, neural methods obtain obvious improvements by introducing the external knowledge (e.g., LOTN and ARGCN) or other solution paradigm (e.g., Dual-MRC and MRC-MVT). Among the neural methods, MRC-MVT achieves very competitive results on all datasets with using machine reading framework. Nevertheless, our MGCR still outperforms it by {1.18\%, 2.01\%, 2.13\%, and 1.63\%} in F1-score respectively on 14res, 15res, 16res and 14lap datasets. These comparisons demonstrate the great superiority of MGCR in exploiting unlabeled data to reduce distribution shift for TOWE, thereby successfully boosting the extraction performance.


\item With few labeled TOWE data, MGCR can also bring great performance gain.
As illustrated in Table~\ref{tab:dataset}, the labeled training sentences of the 14res dataset is about twice the amount of the 15res dataset, while MGCR outperforms MRC-MVT by {1.18\% and 2.01\%} in F1-score respectively on 14res and 15res datasets. The more improvement on 15res proves that MGCR is promising because it perform well even with few labeled sentences.

\end{itemize}

\subsection{Ablation Study}
To evaluate the effectiveness of each component of MGCR, we conduct an ablation study. As shown in Table \ref{tbl:abl}, we observe different levels of performance degradation on the four TOWE datasets when removing the components from MGCR. Specifically, when removing the pre-trained sentiment classifier, the F1 score of MGCR drops from 0.5\% to 1.5\% on all datasets, indicating that using sentiment-attention scores to emphasize latent opinion words helps better filter the noisy unlabeled sentences. After removing either of the coarse-grained and fine-grained filtering processes, the performance declines on all datasets, demonstrating that these two confidence-based thresholding mechanisms alleviate the issue of confirmation bias caused by noisy training signals in consistency regularization.
In addition, we find that removing the consistency regularization (i.e., supervised learning only) worsens the performance most.

\begin{table}[t!]
\resizebox{\linewidth}{!}{
\begin{tabular}{c|ccc|ccc|ccc}
\midrule
\multirow{2}{*}{\diagbox{$\tau$}{$T$}} & \multicolumn{3}{c|}{0.5} & \multicolumn{3}{c|}{0.7} & \multicolumn{3}{c}{0.9} \\ \cline{2-10} 
 & P & R & F1 & P & R & F1 & P & R & F1 \\ \midrule
0.5 & 89.55 & 86.53 & 88.02 & 88.60 & 88.28 & 88.44 & 87.88 & 89.29 & 88.55 \\
0.7 & 89.58 & 87.63 & 88.60 & 87.80 & 88.77 & 88.28 & \textbf{88.65} & \textbf{89.36} & \textbf{89.01} \\
0.9 & 88.64 & 87.83 & 88.23 & 87.65 & 89.28 & 88.45 & 87.73 & 88.35 & 88.03 \\ \midrule
\end{tabular}
}
\caption{Results (\%) of combinations of sentence-level threshold and word-level threshold on 14res. $T$ and $\tau$ represent sentence-level threshold and word-level threshold respectively.}
\label{tbl:thr-14res}
\centering
\end{table}

\begin{table}[t!]
\resizebox{\linewidth}{!}{
\begin{tabular}{c|lll|lll|lll}
\midrule
\multirow{2}{*}{\diagbox{$\tau$}{$T$}} & \multicolumn{3}{c|}{0.5} & \multicolumn{3}{c|}{0.7} & \multicolumn{3}{c}{0.9} \\ \cline{2-10} 
 & \multicolumn{1}{c}{P} & \multicolumn{1}{c}{R} & \multicolumn{1}{c|}{F1} & \multicolumn{1}{c}{P} & \multicolumn{1}{c}{R} & \multicolumn{1}{c|}{F1} & \multicolumn{1}{c}{P} & \multicolumn{1}{c}{R} & \multicolumn{1}{c}{F1} \\ \midrule
0.5 & 81.31 & 82.62 & 81.94 & 81.17 & 83.57 & 82.33 & 82.16 & 84.17 & 83.12 \\
0.7 & 80.13 & 85.39 & 82.62 & 81.84 & 84.04 & 82.92 & \textbf{84.29} & \textbf{83.37} & \textbf{83.80} \\
0.9 & 81.38 & 83.43 & 82.38 & 81.41 & 84.38 & 82.81 & 81.35 & 84.24 & 82.73 \\ \midrule
\end{tabular}
}
\caption{Results (\%) of combinations of sentence-level threshold and word-level threshold on 15res. $T$ and $\tau$ represent sentence-level threshold and word-level threshold respectively.}
\label{tbl:thr-15res}
\centering
\end{table}

\begin{figure}[!t]
    \centering
    \includegraphics[width=0.9\linewidth]{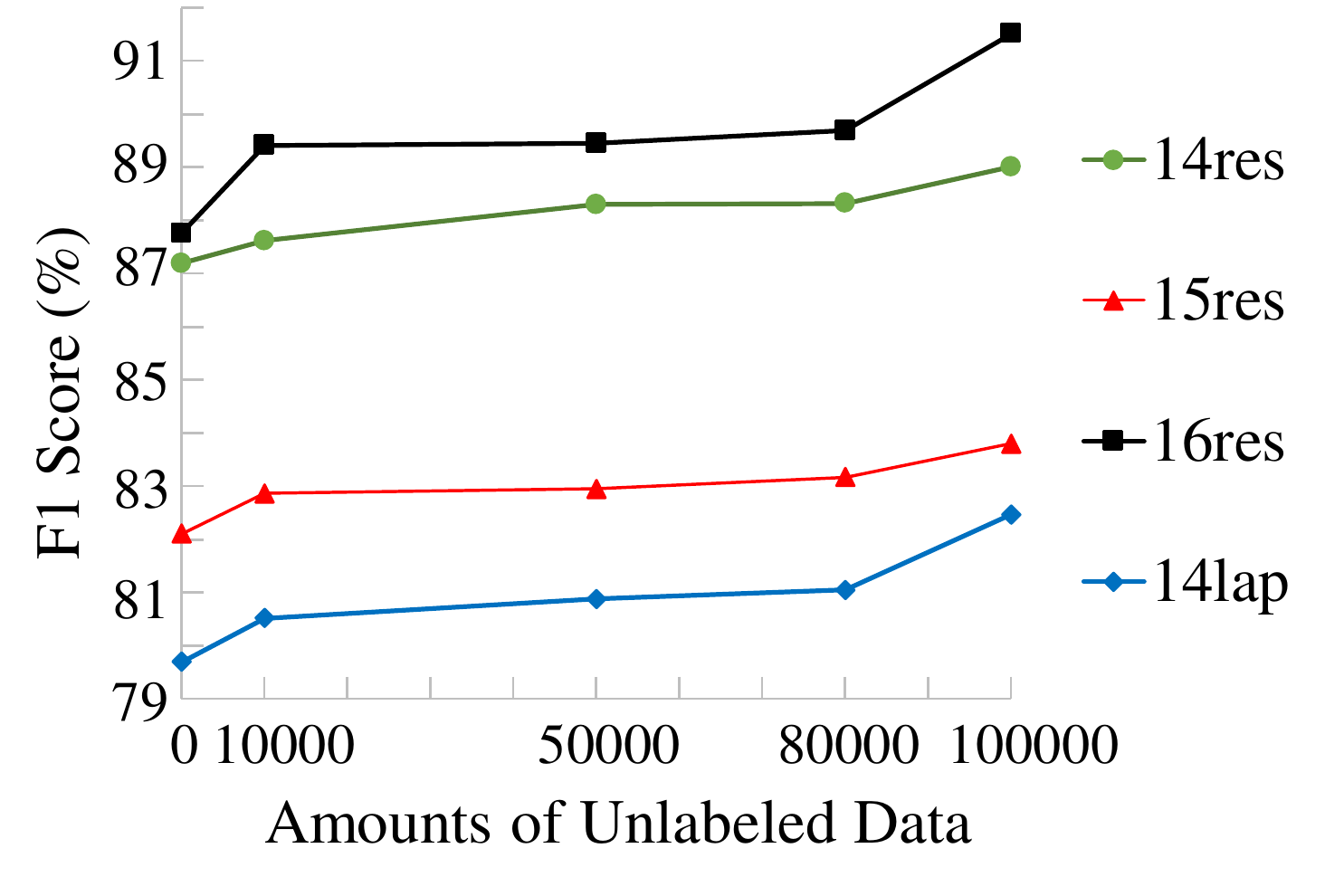}
    \caption{F1-score (\%) on four TOWE datasets with varying amounts of unlabeled data.}
    \label{fig:ulb}
\end{figure}

\subsection{Results of Different Thresholds Combinations }
We present the results of different combinations of sentence-level threshold and word-level thresholds on four TOWE datasets. The range of each threshold is from 0.5 to 0.9. The detailed results for the 14res, 15res, 16res, and 14lap dataset are listed in Tables~\ref{tbl:thr-14res}, \ref{tbl:thr-15res}, \ref{tbl:thr-16res} and \ref{tbl:thr-14lap}, respectively. An interesting finding is that using a high $T$ with a low $\tau$ is the best strategy on most datasets, which means we should filter noisy unlabeled sentences more strictly than filtering noisy unlabeled words.

\subsection{Effect of Amounts of Unlabeled Data}
We conduct experiments by varying the amounts of unlabeled data to investigate the effect of different data amounts. The results are shown in Figure \ref{fig:ulb}. Compared with supervised training only, even a few unlabeled data can improve the performance on all datasets, which validates the effectiveness of utilizing semi-supervised consistency regularization. Besides, the generalization performance is better with more unlabeled data used, because more data effectively reduce the risk of domain shift between test data and training data. This trend demonstrates the potential of MGCR, which are likely to be further improved with even more unlabeled sentences and computing resources.

\begin{table}[t!]
\resizebox{\linewidth}{!}{
\begin{tabular}{c|lll|lll|lll}
\midrule
\multirow{2}{*}{\diagbox{$\tau$}{$T$}} & \multicolumn{3}{c|}{0.5} & \multicolumn{3}{c|}{0.7} & \multicolumn{3}{c}{0.9} \\ \cline{2-10} 
& \multicolumn{1}{c}{P} & \multicolumn{1}{c}{R} & \multicolumn{1}{c|}{F1} & \multicolumn{1}{c}{P} & \multicolumn{1}{c}{R} & \multicolumn{1}{c|}{F1} & \multicolumn{1}{c}{P} & \multicolumn{1}{c}{R} & \multicolumn{1}{c}{F1} \\ \midrule
0.5 & 90.44 & 90.66 & 90.55 & \textbf{91.31} & \textbf{91.74} & \textbf{91.51} & 91.09 & 90.79 & 90.94 \\
0.7 & 90.21 & 90.72 & 90.46 & 89.85 & 91.68 & 90.76 & 90.73 & 90.72 & 90.72 \\
0.9 & 89.45 & 90.35 & 89.88 & 90.16 & 91.36 & 90.75 & 90.90 & 91.36 & 91.13 \\ \midrule
\end{tabular}
}
\caption{Results (\%) of combinations of sentence-level threshold and word-level threshold on 16res. $T$ and $\tau$ represent the sentence-level threshold and word-level threshold respectively.}
\label{tbl:thr-16res}
\centering
\end{table}
\begin{table}[t!]
\resizebox{\linewidth}{!}{
\begin{tabular}{c|lll|lll|lll}
\midrule
\multirow{2}{*}{\diagbox{$\tau$}{$T$}} & \multicolumn{3}{c|}{0.5} & \multicolumn{3}{c|}{0.7} & \multicolumn{3}{c}{0.9} \\ \cline{2-10} 
 & \multicolumn{1}{c}{P} & \multicolumn{1}{c}{R} & \multicolumn{1}{c|}{F1} & \multicolumn{1}{c}{P} & \multicolumn{1}{c}{R} & \multicolumn{1}{c|}{F1} & \multicolumn{1}{c}{P} & \multicolumn{1}{c}{R} & \multicolumn{1}{c}{F1} \\ \midrule
0.5 & 83.29 & 79.65 & 81.42 & 84.85 & 78.07 & 81.32 & 84.43 & 77.51 & 80.82 \\
0.7 & 83.78 & 79.48 & 81.56 & 84.54 & 79.36 & 81.86 & 83.29 & 78.48 & 80.78 \\
0.9 & 82.43 & 78.83 & 80.77 & 84.21 & 78.82 & 81.36 & \textbf{83.76} & \textbf{81.25} & \textbf{82.47} \\ \midrule
\end{tabular}
}
\caption{Results (\%) of combinations of sentence-level threshold and word-level threshold on 14lap. $T$ and $\tau$ represent the sentence-level threshold and word-level threshold respectively.}
\label{tbl:thr-14lap}
\centering
\end{table}

\begin{table*}[!t]
\centering
     \scalebox{0.8}{
        \begin{tabular}{l|c|c|c|c|c}
            \midrule
            Methods                             & NULL & Under-extracted & Over-extracted & Others & Total \\ \midrule
            MGCR                                & \textbf{2}    & \textbf{9}              & \textbf{24}              & 8    & \textbf{43}    \\
            MGCR \textbf{w/o} Pre-trained Sentiment Classifier  & 3    & 12& 29& 11 & 54    \\
            MGCR \textbf{w/o} Filtering Noisy Unlabeled Sentences & 5    & \textbf{9}              & 29              & \textbf{7}      & 55    \\
            MGCR \textbf{w/o} Filtering Noisy Unlabeled Words &4&14&38&11&67\\
            MGCR \textbf{w/o} Consistency Regularization (Labeled Data Only) & 8    & 11              & 44             & 15      & 70    \\ \midrule
        \end{tabular}
    }
    \caption{Statistics of different error types of our MGCR method and different ablation versions on the 16res dataset.} 
      \label{tbl:error}
    \centering
\end{table*}

\subsection{Error Analysis}
We count all the error instances on the 16res dataset to analyze the distribution of different error types. We categorize errors into four types: `NULL': no opinion word is extracted, `Under-extracted': only parts of opinion words are extracted from the input sentences, `Over-extracted': the model extracts redundant opinion words, and `Others': wrong extraction and other error types. As shown in Table \ref{tbl:error}, training with only labeled data suffers from the severe overfitting problem, which leads to many `Over-extracted' errors. For instance, in the sentence ``\emph{open \& cool place with the best pizza and coffee.}'', the overfitted TOWE model extracts both ``\emph{cool}'' and ``\emph{best}'' as opinion words for the opinion target ``\emph{coffee}'', actually the word ``\emph{cool}'' is the opinion word for ``\emph{place}''. Our MGCR reduces the number of `Over-extracted' errors from 44 to 24, indicating that introducing high-quality unlabeled data improves the generalization performance of the TOWE model. It is worth noting that MGCR also reduces 40\% of total errors compared with training with only labeled data.

\section{Related Work}
\subsection{TOWE}
Aspect-based Sentiment Analysis (ABSA) contains a set of various subtasks~\cite{semEval14,liu2015fine,tang2016effective,wang2016recursive,ning2018improving,zhao2020attention,wu2020grid}. \citeauthor{fan2019towe}~\shortcite{fan2019towe} first propose TOWE as a new subtask to expand the ABSA research, which aims to extract opinion words for a given opinion target from a sentence. They employ several LSTM networks and propose a target-fused neural sequence labeling model that achieves promising results on TOWE. Following the idea, many works design advanced multi-head self-attention or multi-view deep attention mechanism to generate target-specific context representation~\cite{feng2021target, zhang2021enhancement}. On this basis, some works incorporate syntactic knowledge~\cite{feng2021target, zhang2021enhancement} or sentiment knowledge~\cite{aaai/lotn} to further improve the performance of TOWE. There are also some other works that transform TOWE into a question answering problem and adopt the framework of machine reading comprehension (MRC) to solve TOWE~\cite{mao2021joint,zhang2021enhancement}, achieving very competitive results. Different from these works, we argue that insufficient labeled data greatly increase the risk of distribution shift for the TOWE task, and thus propose making use of massive unlabeled raw text to reduce the shift risk.

\subsection{Semi-supervised Learning}
 Low-resource learning including semi-supervised learning~\cite{sohn2020fixmatch,zhang2021flexmatch,xie2020unsupervised,wang2022freematch,wangusb,oliver2018realistic} has proven effective in the natural language processing~\cite{peters2017semi,headaptive,cheng2019variational,gururangan2019variational,izmailov2020semi,sintayehu2021named,clark2018semi,ruder2018strong,liu-etal-2022-semi,lu2022rationale,yang2021exploring}, especially ABSA tasks~\cite{marcacini2018cross, xu2019semi, augenstein2018multi,cheng2019variational,li2020seml}. Consistency regularization~\cite{sajjadi2016regularization} is a very popular semi-supervised learning technique, which has been widely applied in various tasks~\cite{berthelot2019mixmatch, berthelot2019remixmatch, sohn2020fixmatch,xie2020unsupervised, chen2020mixtext}. Its core idea is to improve the robustness of the model by minimizing the discrepancy between data and its perturbation.~\citeauthor{miao2020snippext}~\shortcite{miao2020snippext} first applied consistency regularization to aspect sentiment classification task by interpolating embeddings of input sentences. By contrast, we are the first to successfully apply semi-supervised learning consistency regularization to TOWE.

\section{Conclusion}
The TOWE task suffers from the risk of distribution shift which arises from scarce labeled data. In this paper, we propose the novel MGCR method to increase the exposure of the model to varying distribution shifts by exploiting unlabeled data, and naturally the risk can be reduced. In the MGCR method, two different-grained filters, i.e., sentence-level (coarse-grained) and word-level (fine-grained) confidence-based thresholding, are designed to filter noisy sentences and words for the high-quality exposure. To further underline the possible opinion words during learning, we employ a pre-trained sentiment classifier and incorporate sentiment-attention scores into the sentence-level filter. Experimental results indicate that our MGCR method significantly outperforms all other TOWE methods and achieves state-of-the-art performance on four TOWE datasets. The in-depth analysis demonstrate the effectiveness of each component in MGCR. 




\bibliography{anthology,custom}
\bibliographystyle{acl_natbib}

\end{document}